\pdfoutput=1
\documentclass[11pt]{article}
\usepackage{booktabs}

\usepackage[final]{acl}

\usepackage{times}
\usepackage{latexsym}
\usepackage{booktabs} 
\usepackage{hyperref}
\usepackage{array} % required for text wrapping in tables
\usepackage[T1]{fontenc}
\usepackage[utf8]{inputenc}

\usepackage{microtype}

\usepackage{inconsolata}

\usepackage{graphicx}

\title{TIFIN India at SemEval-2025: Harnessing Translation to Overcome Multilingual IR Challenges in Fact-Checked Claim Retrieval}

\author{
  Prasanna Devadiga \\ TIFIN India \\ {\hypersetup{urlcolor=black}\href{mailto:prasanna@askmyfi.com}{prasanna@askmyfi.com}}
  \And
  Arya Suneesh \\ TIFIN India  \\ {\hypersetup{urlcolor=black}\href{mailto:arya.suneesh@askmyfi.com}{arya.suneesh@askmyfi.com}}
  \And
  Pawan Kumar Rajpoot \\ TIFIN India  \\ {\hypersetup{urlcolor=black}\href{mailto:pawan@askmyfi.com}{pawan@askmyfi.com}}
  \AND
  Bharatdeep Hazarika \\ TIFIN India \\{\hypersetup{urlcolor=black}\href{mailto:bharatdeep@askmyfi.com}{bharatdeep@askmyfi.com}}
  \And
  Aditya U Baliga \\ TIFIN India, IIIT Kottayam  \\ {\hypersetup{urlcolor=black}\href{mailto:aditya22bcs54@iiitkottayam.ac.in}{aditya@askmyfi.com}}
}

\begin{document}
\maketitle
\begin{abstract}
We address the challenge of retrieving previously fact-checked claims in monolingual and crosslingual settings - a critical task given the global prevalence of disinformation. Our approach follows a two-stage strategy: a reliable baseline retrieval system using a fine-tuned embedding model and an LLM-based reranker. Our key contribution is demonstrating how LLM-based translation can overcome the hurdles of multilingual information retrieval. Additionally, we focus on ensuring that the bulk of the pipeline can be replicated on a consumer GPU. Our final integrated system achieved a success@10 score of 0.938 ($\sim$0.94) and 0.81025 on the monolingual and crosslingual test sets respectively. The implementation code and trained models are publicly available at our repository\footnote{\url{https://github.com/babel-projekt/semeval_task_7_2025}}.
\end{abstract}

\section{Introduction}
Misinformation poses serious risks to society worldwide, with the World Economic Forum ranking it among the most pressing global threats. Social media acts as a key channel for false content, breaking down trust in institutions and dividing communities, an effect seen most clearly during political events. Our work addresses this problem through SemEval-2025 Shared Task 7, which focuses on developing multilingual retrieval systems that can identify previously fact-checked claims, from the curated MultiClaim dataset, when given social media posts in various languages. Success is measured simply: the system is considered effective if a relevant fact-checked claim appears among the top 10 retrieved results.

\subsection{Task Overview}
SemEval-2025 Shared Task 7 \citep{semeval2025task7} introduces the problem of finding previously fact-checked claims across multiple languages - a task difficult to perform manually, especially when claims and fact-checks appear in different languages. The task is split into monolingual and crosslingual tracks, allowing participants to build systems that help fact-checkers identify relevant fact-checks for social media posts. The task uses a subset of the MultiClaim dataset \citep{pikuliak-etal-2023-multilingual} comprising posts and fact-checks in various language pairs. The dataset provides rich information, including the original post text, text extracted from images via OCR, and translations. Systems are evaluated using the success@10 metric, which measures whether a correct match appears in the top 10 results. This research addresses a real-world challenge faced by fact-checkers who struggle to keep up with misinformation that spreads globally across language boundaries, often requiring knowledge of many languages to identify existing fact-checks effectively.

% \subsection{Contributions and Highlights}
% Our experimental framework encompassed a systematic evaluation of embedding models spanning MPNET, MiniLM, and the Stella series, with model sizes ranging from 90M to 1.5B parameters. To enhance retrieval performance, we augmented the training dataset through LLM-generated summaries and translations, which served as additional training data for the first-stage retrieval model. The implementation of hard-negative mining techniques yielded substantial improvements in embedding model effectiveness. We replaced our initial Cohere reranker with an LLM-based reranking approach for the final candidate selection phase. Through these methodological refinements, our system achieved a success@10 metric of 0.936, resulting in a third-place ranking in the global competition standings.

\section{Related Work}
Research in fake news detection has evolved significantly over the past decade. Early works \citep{10.1145/1963405.1963500, DBLP:journals/corr/abs-1708-01967} introduced methods to assess information credibility on social media, and also explored multi-modal approaches combining textual, user, and network features. Recent works \citep{DBLP:journals/corr/abs-1804-08559, perez-rosas-etal-2018-automatic} focused on linguistic characteristics along with external taxonomies to devise novel detection mechanisms.

Fact verification systems represent a critical approach to combating misinformation. The pioneering work of Thorne et al. \citep{thorne-etal-2018-fever} introduced FEVER (Fact Extraction and VERification), dataset and evaluation framework that has become a benchmark in the field. Building on this foundation, Augenstein et al. \citep{augenstein-etal-2019-multifc} developed multi-domain fact-checking models that can transfer knowledge across different topics and domains.
Neural fact verification approaches have gained prominence, with Nie et al. \citep{DBLP:journals/corr/abs-1811-07039} proposing the Neural Evidence Retriever-Aggregator (NERA) framework, which integrates evidence retrieval with claim verification. Transformer-based architectures have further enhanced verification capabilities, as demonstrated by Wadden et al. \citep{wadden-etal-2020-fact} with their KGAT (Knowledge Graph Attention Network) model that leverages structured knowledge for improved reasoning.

Recent work by \cite{DBLP:journals/corr/abs-2005-11401} introduced retrieval-augmented generation (RAG) models that combine neural retrievers with language models to generate verifiable explanations. 

\section{Methodology}

\begin{figure}
    \centering
    \includegraphics[width=1\linewidth]{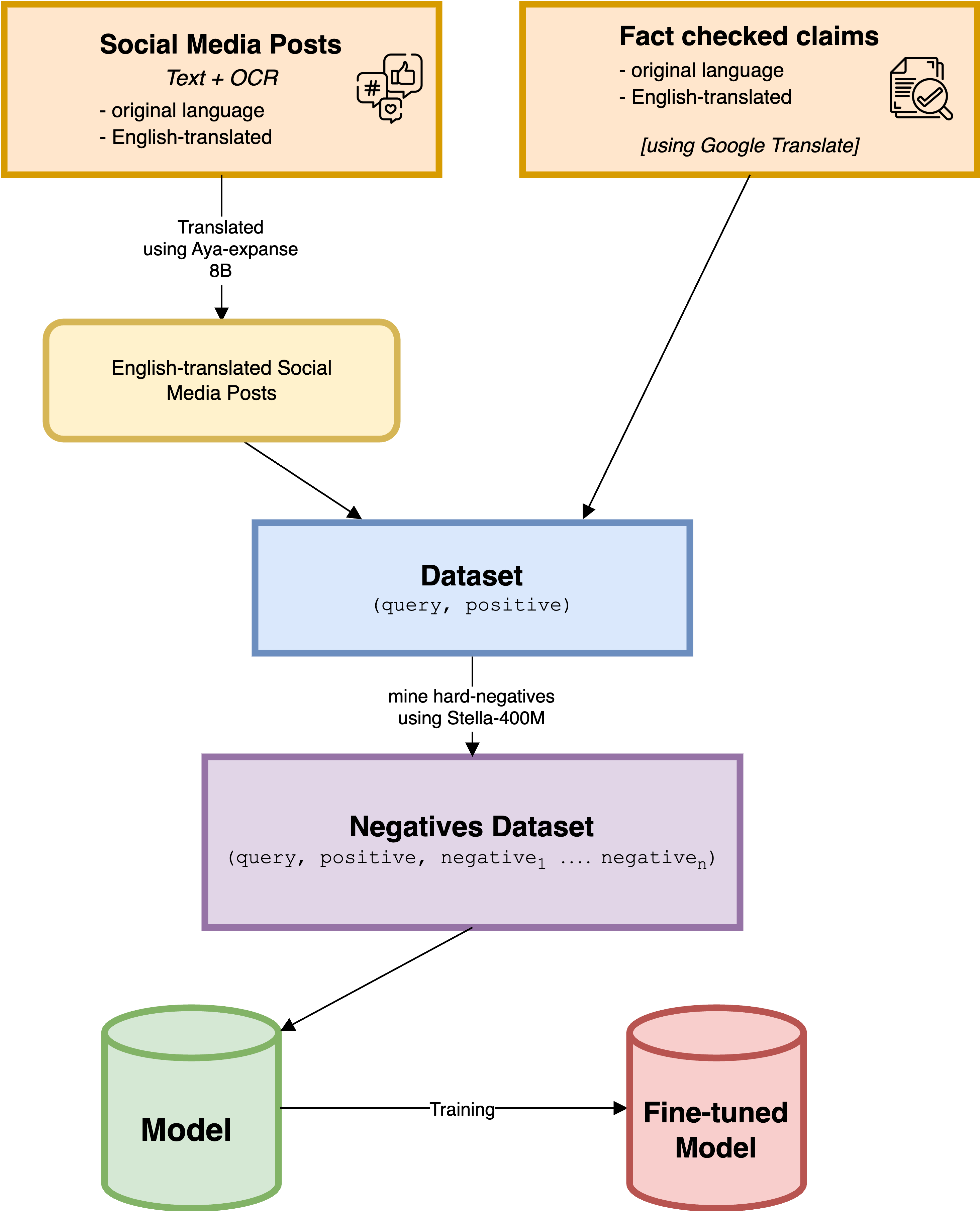}
    \caption{Data Preparation and Model Fine-tuning Pipeline. }
    \label{fig:finetune}
\end{figure}

\subsection{Baseline Embedding-Based Retrieval with Data Augmentation}
Finding relevant fact checks for social media posts in multiple languages is an information retrieval (IR) problem. As a starting point, we opted for an embedding-based retrieval system due to its straightforward implementation and scalability. While the dataset includes social media posts and fact checks in their original languages, it also provides English translations via Google Translate. These translations proved particularly useful, allowing us to begin with a purely English retrieval setup while deferring the complexities of multilingual IR—challenges we explore in detail in a later section.

To establish a strong reference model, we selected the top 10 performing models from the MTEB English v2 benchmark\hyperref[]{} from MTEB English v2 \citep{muennighoff2022mteb} considering only those with fewer than 1B parameters to ensure compatibility with consumer GPUs. This choice was motivated by our decision to initially focus on English translations, ensuring a more controlled evaluation before tackling cross-lingual retrieval challenges. To complement these selections, we included MPNet v2 (built on top of MPNet) \citep{song2020mpnet} and the GTR-T5 \citep{ni2021large} family, as they were among the top contenders in the MultiClaim dataset paper. We also include MiniLM \citep{wang2020minilm}, a 90M parameter model, to better quantify performance trade-offs across model sizes

Ultimately, we chose Stella 400M \citep{zhang2024jasper} due to its strong baseline performance as described in \hyperref[tab:model_comparison]{this table} for our embedding purposes. 

\subsection{Data Augmentation}

Multilingual IR presents significant challenges due to diverse writing systems, grammatical structures, and computational constraints for low-resource languages. The MultiClaim dataset, with social media posts in 27 languages and fact-checking articles in 39 languages, exemplifies these complexities. Our solution was to translate all social media posts into English, simplifying system design by eliminating the need for language-specific models while reducing computational demands. This approach solves memory and processing constraints of multilingual systems. While translation may cause loss of meaning, it leverages advanced English models to improve accuracy, enabling cross-lingual pattern learning that particularly benefits low-resource languages.

For translation, we selected the Aya Expanse 8B model (\hyperref[tab:reasoning_aug_example]{Prompt}) \citep{dang2024ayaexpansecombiningresearch} over Google Translate based on compelling empirical evidence. Recent studies demonstrate that large language models produce translations that are 4-18\% more accurate than traditional systems according to BLEU scores, with superior capability in preserving contextual nuances and idiomatic expressions. \footnote{\url{https://medium.com/@flavienb/machine-translation-in-2023-48e14eb4cb71}} Aya's coverage of 23 languages outperforms comparable models like mT0 and Bloomz \citep{muennighoff2022crosslingual} in benchmarks, making it ideal for our diverse dataset. This selection is further supported by research showing that newer LLM-based translation systems require fewer post-translation corrections and achieve better performance compared to conventional approaches, including Google's PaLM 2 model, \citep{anil2023palm} outperforms Google Translate.

The single-model translation approach offers considerable benefits beyond computational efficiency. It simplifies development, deployment, and maintenance relative to managing multiple specialized models while enabling transfer learning where cross-lingual patterns benefit all languages, especially those with limited training data. This design choice to use translation was further inspired by Sarvam's IndicSuite \citep{khan2024indicllmsuite}  system, which achieved state-of-the-art results for Indian languages through \cite{gala2023indictrans} machine translation of large datasets. Our approach balances quality and efficiency by converting languages into English while preserving semantic meaning, addressing challenges of different writing systems and grammar structures, while utilizing English-focused embedding models for better retrieval results.

\subsection{Re-ranking strategies}
In our exploration of re-ranking methodologies to further improve upon the initial retrieval performance achieved through the Stella 400M embedding model (which attained 0.86 average success@10 on the development set), we investigated several approaches including cross encoders, Colbert v2 \citep{10.1145/3397271.3401075, santhanam-etal-2022-colbertv2}, T5/Seq2Seq \citep{raffel2020exploring} architectures, and a Cohere reranker. 
Although initial experiments with the Cohere reranker \footnote{https://cohere.com/blog/rerank} showed promising improvements, further analysis revealed that similar performance gains could be achieved by fine-tuning the embedding model with hard negatives as described in Section 3.4, effectively neutralizing its advantages. 
Based on these findings, we transitioned to an LLM-based re-ranking strategy that directly processes the top 50 candidates, evaluating several large language models including Gemini 1.5 Pro \cite{team2024gemini}, Meta Llama 3.1 70B, Llama 3.3 70B Instruct \citep{dubey2024llama}, and Qwen 2.5 72B Instruct \citep{yang2024qwen2}. Qwen 2.5 72B Instruct demonstrated superior performance and was ultimately selected for our final system (\hyperref[tab:reranking_prompt]{Prompt}). This approach leverages the model's world knowledge and semantic understanding to establish the final ranking order, yielding additional performance improvements in our system's overall effectiveness. 

\begin{figure}
    \centering
    \includegraphics[width=1\linewidth]{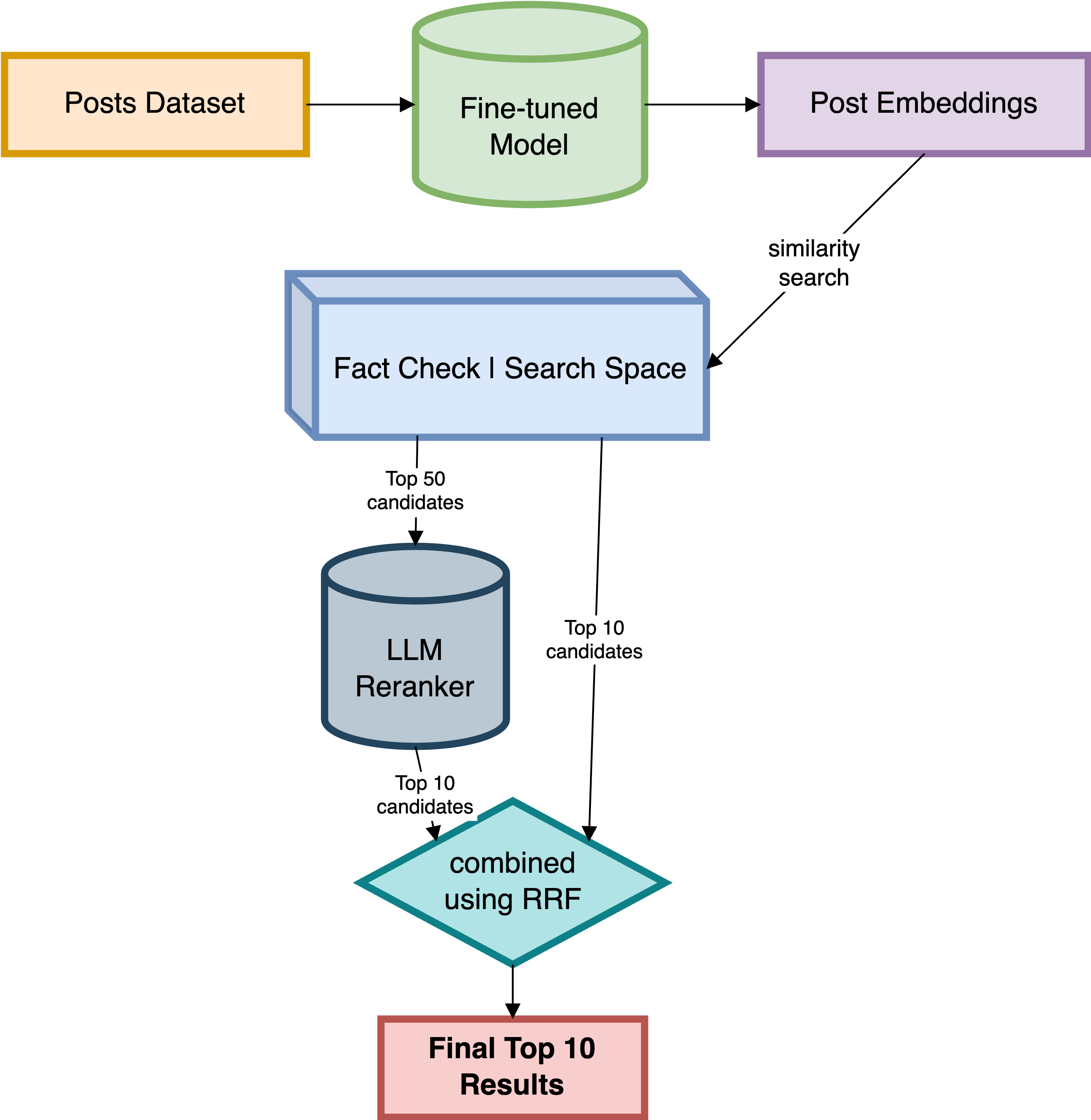}
    \caption{Two-Stage Retrieval and Ranking Architecture}
    \label{fig:enter-label}
\end{figure}

\subsection{Hard-Negative Mining and Finetuning of the Embedding Models}
The effectiveness of re-ranking is inherently tied to the quality of the initial candidate set, necessitating improvements to the underlying embedding model. To this end, we investigated fine-tuning strategies through extensive experimentation, comparing two approaches.
The first approach relied solely on positive query-document pairs, but this provided only marginal improvements in retrieval performance, suggesting that learning from relevant matches alone was insufficient. In contrast, incorporating hard negatives into the training process led to substantial gains.
We leveraged Sentence Transformers' \footnote{https://www.sbert.net/index.html} native negative mining capabilities, systematically varying the number of negative examples per query from 5 to 80. Our analysis revealed that performance gains plateaued at approximately 40 negatives per query. To optimize computational efficiency while maintaining performance benefits, we ultimately selected 20 negatives per query, which successfully elevated our system's success@10 metric beyond 0.90.

\begin{table*}
  \centering
  \resizebox{\textwidth}{!}{%
  \begin{tabular}{lccccccccccc}
    \hline
    \textbf{Model} & \textbf{pol S@10} & \textbf{eng S@10} & \textbf{msa S@10} & \textbf{por S@10} & \textbf{deu S@10} & \textbf{ara S@10} & \textbf{spa S@10} & \textbf{fra S@10} & \textbf{tha S@10} & \textbf{tur S@10} & \textbf{avg S@10} \\
    \hline
    NovaSearch/stella\_en\_400M\_v5 & \textbf{0.788} & \textbf{0.678} & 0.956 & 0.658 & 0.774 & \textbf{0.914} & \textbf{0.75} & \textbf{0.876} & \textbf{0.978} & 0.786 & \textbf{0.815} \\
    sentence-transformers/all-MiniLM-L6-v2 & 0.726 & 0.61 & 0.956 & 0.6 & 0.71 & 0.864 & 0.66 & 0.834 & 0.956 & 0.754 & 0.767 \\
    sentence-transformers/all-mpnet-base-v2 & 0.674 & 0.602 & 0.956 & 0.552 & 0.67 & 0.842 & 0.614 & 0.804 & 0.934 & 0.692 & 0.734 \\
    BAAI/bge-base-en & 0.694 & 0.614 & 0.935 & 0.53 & 0.686 & 0.84 & 0.624 & 0.826 & 0.939 & 0.726 & 0.741 \\
    BAAI/bge-large-en & 0.674 & 0.628 & 0.956 & 0.55 & 0.678 & 0.862 & 0.618 & 0.832 & 0.923 & 0.688 & 0.741 \\
    avsolatorio/GIST-Embedding-v0 & 0.756 & 0.65 & 0.978 & 0.626 & 0.752 & 0.898 & 0.72 & 0.858 & 0.950 & 0.774 & 0.796 \\
    avsolatorio/GIST-large-Embedding-v0 & 0.766 & 0.664 & 0.956 & 0.648 & 0.764 & 0.904 & 0.722 & 0.866 & 0.967 & 0.788 & 0.804 \\
    avsolatorio/GIST-small-Embedding-v0 & 0.73 & 0.624 & \textbf{1} & 0.608 & 0.726 & 0.866 & 0.656 & 0.852 & 0.961 & 0.774 & 0.779 \\
    thenlper/gte-large & 0.768 & 0.656 & 0.956 & 0.642 & \textbf{0.782} & 0.906 & 0.722 & 0.874 & 0.95 & 0.796 & 0.805 \\
    sentence-transformers/gtr-t5-large & 0.752 & 0.662 & 0.956 & 0.652 & 0.76 & 0.882 & 0.694 & 0.866 & 0.934 & 0.79 & 0.794 \\
    sentence-transformers/gtr-t5-xl & 0.76 & 0.658 & 0.967 & 0.646 & 0.778 & 0.884 & 0.68 & 0.868 & 0.95 & \textbf{0.798} & 0.799 \\
    intfloat/multilingual-e5-large & \textbf{0.788} & 0.666 & 0.956 & \textbf{0.66} & 0.76 & 0.91 & 0.72 & 0.858 & 0.95 & 0.758 & 0.802 \\
    mixedbread-ai/mxbai-embed-large-v1 & 0.71 & 0.646 & 0.956 & 0.588 & 0.74 & 0.884 & 0.688 & 0.848 & 0.961 & 0.76 & 0.778 \\
    sentence-transformers/paraphrase-multilingual-mpnet-base-v2 & 0.714 & 0.584 & 0.924 & 0.568 & 0.698 & 0.842 & 0.598 & 0.808 & 0.939 & 0.734 & 0.741 \\
    WhereIsAI/UAE-Large-V1 & 0.708 & 0.638 & 0.956 & 0.59 & 0.736 & 0.886 & 0.682 & 0.856 & 0.95 & 0.752 & 0.775 \\
    \hline
  \end{tabular}%
  }
  \caption{\label{tab:model_comparison} Comparison of different base models on S@10 metric across languages.}
\end{table*}

\begin{table*}
  \centering
  \small
  \resizebox{\textwidth}{!}{%
  \begin{tabular}{lcccccccccc}
    \hline
    \textbf{S@10 (avg)} & \textbf{S@10 (eng)} & \textbf{S@10 (fra)} & \textbf{S@10 (deu)} & \textbf{S@10 (por)} & \textbf{S@10 (spa)} & \textbf{S@10 (tha)} & \textbf{S@10 (msa)} & \textbf{S@10 (ara)} & \textbf{S@10 (tur)} & \textbf{S@10 (pol)} \\
    \hline
    0.9383 & 0.880 & 0.954 & 0.936 & 0.902 & 0.960 & 0.9945 & 1 & 0.966 & 0.904 & 0.886 \\
    \hline
  \end{tabular}%
  }
  \caption{S@10 performance of TIFIN India across languages.}
  \label{tab:s10_languages}
\end{table*}

\subsection{Hybrid Search}
Integrating a sparse embedding model along with a dense embedding model represents a common configuration employed within RAG systems for enhanced reliability. Performance typically improves because sparse models excel at capturing exact lexical matches and relevant keywords, while dense models focus on semantic features and contextual understanding. However, in our experiments, we observed that the use of BM25 alongside base embedding models improved performance only for smaller models, such as MiniLM, but appeared to reduce the effectiveness of top-performing embedding models, such as Stella.
This performance degradation likely occurs because sophisticated embedding models already effectively encode both lexical and semantic information within their high-dimensional representations, making the additional signal from sparse retrieval redundant or potentially introducing noise that dilutes the precision of the embedding model. An additional observation from the MultiClaim dataset study  demonstrated that BM25 exhibits a higher false positive rate as the fact-check pool size increases, particularly affecting languages with larger collections \citep{pikuliak-etal-2023-multilingual}. Consequently, rather than combining sparse and dense approaches, we evaluated using multiple dense models within a hybrid search and established final rankings using Reciprocal Rank Fusion (RRF) \citep{cormack2009reciprocal}. The simplest configuration that yielded performance improvements was combining the model's base retrieval results with its reranked outputs using RRF.

\subsection{Final Pipeline}
Our final system implements a two-step pipeline, with the first stage employing a fine-tuned Stella 400M trained on our augmented dataset to perform baseline retrieval. This is followed by a Qwen-based reranker that processes the top 50 candidates. The pipeline culminates in a hybrid search mechanism using RRF to combine results from the fine-tuned model and reranked outputs. This integrated approach delivers robust retrieval performance while maintaining practical computational requirements for real-world applications.

\subsection{Compute Requirements}
We ran timing experiments to estimate our approach's compute requirements. As shown in Table~\ref{tab:compute_time}, translating the test-set posts (monolingual and cross-lingual) took 61 minutes, and fine-tuning the Stella embedding model took 35 minutes—both on an NVIDIA RTX 4090. These numbers show that the core system can run on consumer hardware without requiring large-scale infrastructure.

\begin{table}[h]
  \centering
  \small
  \begin{tabular}{p{4.5cm} r}
    \hline
    \textbf{Task} & \textbf{Time (minutes)} \\
    \hline
    Translation (monolingual + crosslingual posts) & 61 \\
    Finetuning Stella embedding model & 35 \\
    \hline
  \end{tabular}
  \caption{Compute time for key components (test-set only), measured on an NVIDIA RTX 4090}
  \label{tab:compute_time}
\end{table}

\section{Results}
Our experimental evaluation revealed several key findings about the effectiveness of different retrieval approaches.  

As shown in \hyperref[tab:model_comparison]{Table 1}, the Stella 400M model demonstrated strong performance across multiple languages, achieving a baseline success@10 score of 0.8159. While most models performed well for Malay (msa) and Thai (tha), performance varied considerably for others, particularly German (deu) and Portuguese (por).  

Notably, Stella’s built-in s2p prompt provided a meaningful boost, improving performance from 0.8159 to 0.8305 (+1.8\%), demonstrating its effectiveness in enhancing query formulation. Further, incorporating translated posts yielded a much larger improvement, increasing success@10 to 0.8837 (+6.8\% relative to baseline), highlighting the value of cross-lingual augmentation.  

Fine-tuning on translated posts further enhanced retrieval effectiveness, reaching a success@10 score of 0.9217. This represents an ~11\% relative improvement over the base Stella model and an additional 4.3\% gain over using translated posts alone.  

As shown in \hyperref[tab:reranker_delta]{Table 4}, conventional rerankers generally degraded performance, with deltas ranging from -0.127 to -0.032. However, the Qwen2.5-72B-Instruct reranker provided a modest but positive impact (+0.025).  

Our final configuration (\hyperref[tab:config_performance]{Table 5}), which combined fine-tuning with translation, reranking, and reciprocal rank fusion (RRF), achieved the best overall performance at 0.938. However, the improvement from reranking and RRF was relatively modest (+1.7\%) compared to the gains from fine-tuning and translation, suggesting that most of the effectiveness stemmed from improving the retrieval model rather than post-processing.

\begin{table}[h]
  \centering
  \small
  \begin{tabular}{>{\raggedright\arraybackslash}p{0.6cm}lr}
    \hline
    \textbf{Base} & \textbf{Reranker} & \textbf{S@10 $\Delta$} \\
    \hline
    {minilm} 
    & colbert-ir/colbertv2.0 & -0.032 \\
    & mixedbreath-ai/mbai-rerank-large-v1 & -0.065 \\
    & mixedbread-ai/mxbai-rerank-base-v1 & -0.081 \\
    & unicamp-dl/InRanker-base & -0.037 \\
    \hline
    {stella} 
    & colbert-ir/colbertv2.0 & -0.071 \\
    & mixedbreath-ai/mbai-rerank-large-v1 & -0.105 \\
    & mixedbread-ai/mxbai-rerank-base-v1 & -0.127 \\
    & unicamp-dl/InRanker-base & -0.075 \\
    & \textbf{Qwen/Qwen2-72B-Instruct} & \textbf{+0.025} \\
    \hline
  \end{tabular}
  \caption{\label{tab:reranker_delta} Performance delta (S@10) of various rerankers compared to base models.\protect\footnotemark}
\end{table}
\footnotetext{Due to computational budget constraints, Qwen re-ranking was not evaluated with the MiniLM retriever.}

\begin{table}[h]
  \centering
  \small
  % Adjust width as needed (e.g., 4.5cm or 5.2cm).
  \begin{tabular}{p{4cm} r}
    \hline
    \textbf{Config} & \textbf{Performance (S@10)} \\
    \hline
    Stella & 0.8159 \\
    Stella + Prompt (s2p query) & 0.8305 \\
    Stella + Translated Posts & 0.8837 \\
    Finetuned Stella + Translated Posts & 0.9217 \\
    \textbf{Finetuned Stella + Translated Posts + Reranking + RRF} & \textbf{0.938} \\
    \hline
  \end{tabular}
  \caption{Performance (S@10) for various system configurations on the monolingual test set}
  \label{tab:config_performance}
\end{table}

\section*{Limitations}
Our research operated with defined hardware limitations, using an NVIDIA RTX 4090 GPU with 24GB VRAM and 64GB system memory while leveraging TogetherAI's \footnote{https://www.together.ai/} cloud services for a serverless LLM endpoint. These resource limitations significantly influenced our evaluation scope, restricting our ability to comprehensively assess models exceeding 1 billion parameters, despite evidence suggesting such larger models have established state-of-the-art performance benchmarks. These practical constraints represent important context for interpreting our experimental results and methodology.

\section{Conclusion and Future Work}
We developed a novel approach for retrieving fact-checked claims across multiple languages using a multi-stage system that combines LLM-based translation, embedding models improved through hard-negative mining, and an advanced reranking approach incorporating reciprocal rank fusion. Our results on the task data set show that this integrated approach works well to find relevant information across language boundaries, achieving competitive performance. Additionally , we show that LLM-based re-ranking is not required from a  performance standpoint as it offers minimal gains. Our recommendation is to simply use translation along with a finetuned embedding model, as this combination captures most of the performance gains.
Future work includes testing generalizability on other datasets, applying these techniques to broader information retrieval problems, evaluating quantized embedding vectors for improved efficiency, and exploring LLM-based translation approaches for low-resource languages. 

\bibliography{acl_latex}

\appendix

\section{Appendix}
\label{sec:appendix}

\begin{table}[h]
    \centering
    \small
    \caption{Prompt used for LLM-based translation}
    \begin{tabular}{@{}p{1\linewidth}@{}}
    \toprule
    You are given text (possibly noisy social media data) that may be partially or entirely in a non-English language. 
    It could contain repeated emojis, excessive punctuation, or minor errors.\\
    Your task is to produce a “cleaned but faithful” English version. Specifically:\\
    1) If the text is not in English, translate it to English as literally as possible.\\
    2) Preserve important meaning, tone, and references (e.g., named entities, hashtags, or domain-specific terms).\\
    3) Remove or reduce meaningless filler (like repeated punctuation or stray symbols) without losing factual content.\\
    4) Avoid adding your own commentary, opinions, or extra interpretation. Keep the style and intent aligned with the original.\\
    \bottomrule
    \end{tabular}
    \label{tab:reasoning_aug_example}
\end{table}

\begin{table}[h]
    \centering
    \small
    \caption{Prompt used for LLM-based re-ranking}
    \begin{tabular}{@{}p{1\linewidth}@{}}
    \toprule

\#\# You are an expert fact-checker and information retrieval specialist. Your task is to analyze a query and a set of articles to identify the most relevant ones for fact-checking purposes.\\
\\
\#\# Task:\\
1. Review the query that needs fact-checking\\
2. Analyze the candidate articles provided\\
3. Select the 10 most relevant articles that would be most useful for fact-checking the query\\
4. Return ONLY the article IDs of these 10 articles in a tab-separated format\\
\\
\#\# Important Instructions:\\
- Focus on selecting articles that:\\
  * Directly address the claim in the query\\
  * Provide factual evidence or counter-evidence\\
  * Come from reliable sources\\
  * Contain specific details relevant to the query\\
  * Cover different aspects of the claim for comprehensive fact-checking\\
- Output format must be EXACTLY:\\
  * Only article IDs\\
  * Tab-separated\\
  * One line only\\
  * Top 10 articles in order of relevance\\
  * No explanations or additional text\\
\\
\#\# Query for fact-checking: ***QUERY*** \\
\#\# Data Augmentations:  ***AUGMENTATION*** \\ 
\#\# Candidate Articles:\\ 
***ARTICLE 1***\\
***ARTICLE 2*** \\
- \\
- \\
- \\
***ARTICLE N*** \\
ONLY RETURN tab-seperated IDs....NOTHING ELSE\\
    \bottomrule
    \end{tabular}
    \label{tab:reranking_prompt}
\end{table}

\end{document}